\documentclass[a4paper,conference]{IEEEtran}
\ifCLASSINFOpdf
\else
\fi
\makeatletter
\DeclareRobustCommand\onedot{\futurelet\@let@token\@onedot}
\def\@onedot{\ifx\@let@token.\else.\null\fi\xspace}

\def\eg{\emph{e.g}\onedot} 
\def\ie{\emph{i.e}\onedot} 
\def\cf{\emph{c.f}\onedot}

\makeatother

\usepackage{amsfonts}
\usepackage{amssymb}
\usepackage{booktabs}
\usepackage{comment}
\usepackage{graphicx}
\usepackage{multirow}
\usepackage{xspace}
\usepackage{amsmath}
\usepackage{bbm}

\usepackage{bbold}
\usepackage{mathtools}

\begin{document}
%
% paper title
% Titles are generally capitalized except for words such as a, an, and, as,
% at, but, by, for, in, nor, of, on, or, the, to and up, which are usually
% not capitalized unless they are the first or last word of the title.
% Linebreaks \\ can be used within to get better formatting as desired.
% Do not put math or special symbols in the title.
\title{CobNet: Cross Attention on Object and Background for Few-Shot Segmentation}

% author names and affiliations
% use a multiple column layout for up to three different
% affiliations
\author{\IEEEauthorblockN{Haoyan Guan}
\IEEEauthorblockA{Department of Informatics\\
 King's College London\\
 London, UK\\
 Email: haoyan.guan@kcl.ac.uk}
\and
\IEEEauthorblockN{Michael Spratling}
\IEEEauthorblockA{Department of Informatics\\
 King's College London\\
 London, UK\\
 Email: michael.spratling@kcl.ac.uk}}

% conference papers do not typically use \thanks and this command
% is locked out in conference mode. If really needed, such as for
% the acknowledgment of grants, issue a \IEEEoverridecommandlockouts
% after \documentclass

% for over three affiliations, or if they all won't fit within the width
% of the page, use this alternative format:
%
%\author{\IEEEauthorblockN{Michael Shell\IEEEauthorrefmark{1},
%Homer Simpson\IEEEauthorrefmark{2},
%James Kirk\IEEEauthorrefmark{3},
%Montgomery Scott\IEEEauthorrefmark{3} and
%Eldon Tyrell\IEEEauthorrefmark{4}}
%\IEEEauthorblockA{\IEEEauthorrefmark{1}School of Electrical and Computer Engineering\\
%Georgia Institute of Technology,
%Atlanta, Georgia 30332--0250\\ Email: see http://www.michaelshell.org/contact.html}
%\IEEEauthorblockA{\IEEEauthorrefmark{2}Twentieth Century Fox, Springfield, USA\\
%Email: homer@thesimpsons.com}
%\IEEEauthorblockA{\IEEEauthorrefmark{3}Starfleet Academy, San Francisco, California 96678-2391\\
%Telephone: (800) 555--1212, Fax: (888) 555--1212}
%\IEEEauthorblockA{\IEEEauthorrefmark{4}Tyrell Inc., 123 Replicant Street, Los Angeles, California 90210--4321}}

% use for special paper notices
%\IEEEspecialpapernotice{(Invited Paper)}

% make the title area
\maketitle

% As a general rule, do not put math, special symbols or citations
% in the abstract
\begin{abstract}
Few-shot segmentation aims to segment images containing objects from previously unseen classes using only a few annotated samples. Most current methods focus on using object information extracted, with the aid of human annotations, from support images to identify the same objects in new query images. However, background information can also be useful to distinguish objects from their surroundings. Hence, some previous methods also extract background information from the support images. In this paper, we argue that such information is of limited utility, as the background in different images can vary widely. To overcome this issue, we propose CobNet which utilises information about the background that is extracted from the query images without annotations of those images. Experiments show that our method achieves a mean Intersection-over-Union score of 61.4\% and 37.8\% for 1-shot segmentation on PASCAL-$5^i$ and COCO-$20^i$ respectively, outperforming previous methods. It is also shown to produce state-of-the-art performances of 53.7\% for weakly-supervised few-shot segmentation, where no annotations are provided for the support images. \end{abstract}

% no keywords

% For peer review papers, you can put extra information on the cover
% page as needed:
% \ifCLASSOPTIONpeerreview
% \begin{center} \bfseries EDICS Category: 3-BBND \end{center}
% \fi
%
% For peerreview papers, this IEEEtran command inserts a page break and
% creates the second title. It will be ignored for other modes.
\IEEEpeerreviewmaketitle

\section{Introduction}
\label{sec_intro}

Artificial intelligence has enjoyed remarkable success in recent years, and achieved extraordinary results in computer vision \cite{he2016deep,redmon2016you,long2015fully}. These results are based on large labeled datasets which are used to fit nonlinear functions using human-generated scene descriptions. Collecting such large-scale datasets is time-consuming, and this approach seems to lack the intelligence shown by humans, who can accurately identify the category of an object in a new image having only seen one, or a few, examples of this new category. In order to avoid the need to large annotated datasets, few-shot learning (FSL) attempts to learn new concepts from few examples. This paper is specifically concerned with few-shot segmentation (FSS) which aims to train a segmentation model on a novel category using a few images and their corresponding annotated segmentation masks. The number of labeled images in this support set is typically either 1 or 5, giving rise to two sub-tasks used to evaluate FSS referred to as 1-shot and 5-shot segmentation.
%Different sub-tasks within FSS are described as m-shot n-way, where m is the size of the support set, and n is the number of object classes. 

Most current FSS algorithms \cite{zhang2019canet,siam2019amp,zhang2019pyramid,yang2020prototype,boudiaf2020few} employ a similar sequence of operations. Firstly, feature extraction is typically performed using a CNN that has been pre-trained on ImageNet \cite{russakovsky2015imagenet,yang2020prototype,siam2020weakly,zhang2019canet}. 
The output of certain convolution layers is used to represent an input image in deep feature space, and the process of extracting features from the support images and the query image is identical. 
Secondly, the support masks are used to identify features extracted from the support images which represent the foreground objects. Usually, each object class is represented by a single prototype feature vector \cite{wang2019panet,yang2020prototype,tian2020prior}. Finally, the feature vectors extracted from the query image are classified by comparing their similarity with the prototype feature vectors extracted from the support set. The result is a prediction of the segmentation mask for the query image. This predicted segmentation mask is compared to a ground-truth mask and the result of this comparison is used during training to  tune the parameters of the model, and during testing to evaluate performance.
%The parameters used in all stages of the model are tuned by comparing this predicted mask with the human annotated masks provided for the support set images.

\begin{figure}[t]
\centering
\includegraphics[width=\linewidth]{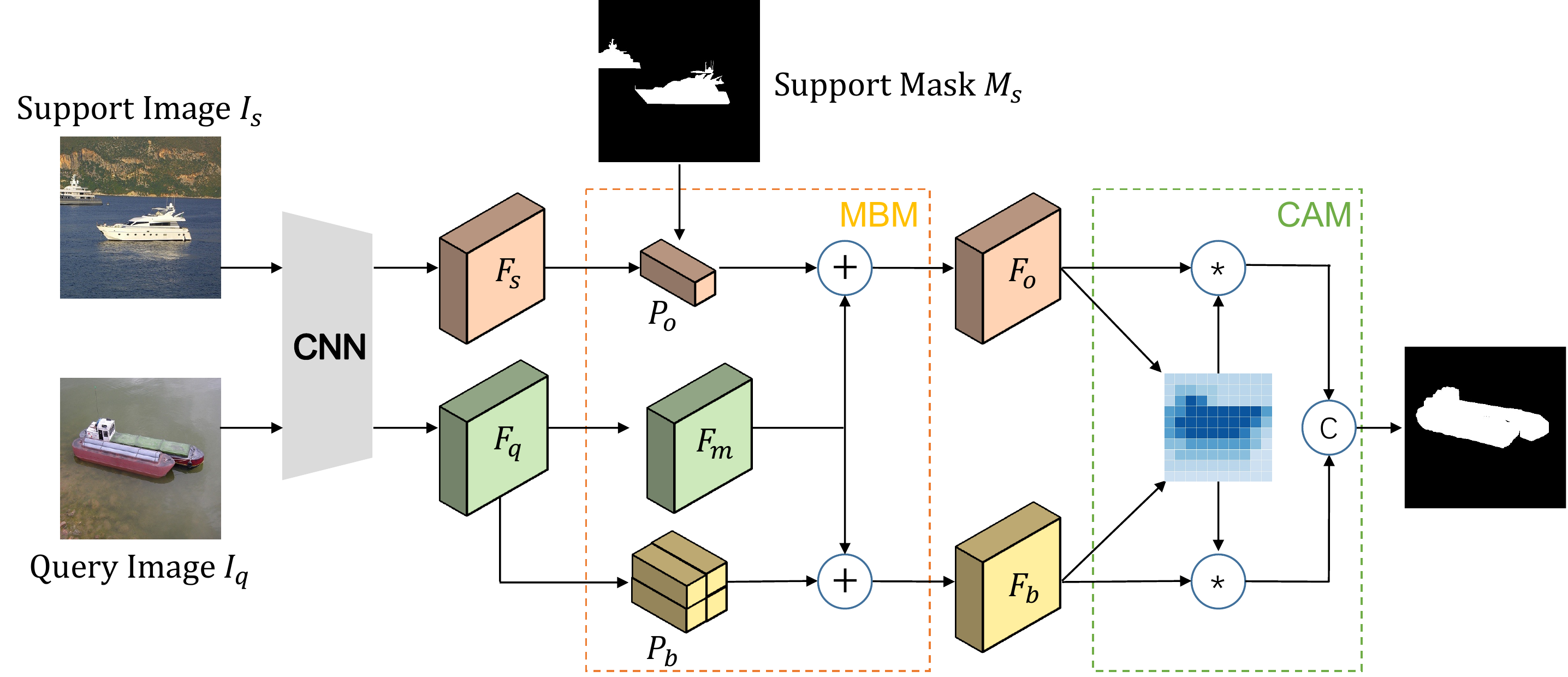}
\caption{An overview of the proposed CobNet architecture in the 1-shot segmentation scenario. $F_{s}$ and $F_{q}$ are features, extracted by a pre-trained CNN, from the support and query images. The MBM (Multi-scale Background Module) obtains an object prototype ($P_{o}$) and background prototypes ($P_{b}$) from the support and query features. The extraction of background information from the query image is the main innovation of our approach, and it enables the MBM to generate features ($F_{o}$ and $F_{b}$ which respectively represent object and background) that are more relevant for segmenting the query image. The CAM (Cross Attention Module) is used to calculate an attention matrix to weight object and background features, and predict the final result.}
\label{fig_overview}
\end{figure}

%A common architecture for FSS is a two-branch network, where the support branch obtains prototypes and the query branch establishes the relationship between prototypes and query features by their similarity. 
Most FSS methods \cite{zhang2019canet,yang2020brinet,tian2020prior,li2021adaptive} only obtain prototypes of objects, ignoring the background information. However, the background provides negative samples that can be mined to reduce false-positives, and hence, make the model more discriminative. Some FSS methods \cite{yang2020prototype,boudiaf2020few,wang2019panet} obtain background prototypes from the support set, then calculate the similarity between background prototypes and query features. However, these methods ignore the fact that the same category object may appear against different backgrounds in different images. For example, a human might appear against a background that is an indoor scene in the support images, but in the context of an outdoor scene in the query image. In such cases, the background prototypes obtained from the support images are unlikely to be helpful in segmenting the query image.

In this paper, we propose a novel framework named CobNet to fix this issue. We argue that the most relevant background information for an image can be obtained from that image itself. Therefore, we extract background prototypes from the query image (the image that is currently being segmented) without the need for a segmentation mask. In contrast, the object prototypes are extracted from the support set by making use of support masks. As masks are unavailable for the query set images, an unsupervised method is necessary to obtain the background prototypes of the query image. Because all unsupervised methods inevitably introduce noise and we are limited by computational complexity, CobNet chooses average pooling (AP) as a simple, but surprisingly effective, unsupervised method to extract background prototypes. AP is also used to obtain object prototypes from the support image. However, in this case, the average pooling is restricted to the region of the support image containing the object by making use of the support mask. As detailed in section~\ref{sec_MBM}, object (background) prototypes are fused with query features to get object (background) features. A cross attention module then calculates an attention weight to balance the relationship between object and background features (more details in section~\ref{sec_CAM}). An overview of the proposed framework is shown in figure \ref{fig_overview}. The main contributions of this work are as follows:

\begin{itemize}
  \item[1)] 
  We develop CobNet which takes a unique approach to FSS by performing negative mining on the query image, so that information about the background is appropriate to the image being segmented.
  %that combines two new modules, MBM and CAM. MBM extracts foreground object prototypes from the support images, and multi-scale background features from the query image. CAM generates an attention weight to balance object and background features.
  \item[2)]
  Experiments show that CobNet achieves a mean Intersection-over-Union (mIoU) score of 61.4\% and 37.8\% for 1-shot segmentation on the PASCAL-$5^i$ and COCO-$20^i$ datasets, outperforming previous approaches to set a new state-of-the-art on standard benchmarks used to assess FSS.
  \item[3)]
  Experiments on PASCAL-$5^i$ show that our method achieves a mIoU score of 53.7\% in the situation where no masks are provided to identify the foreground objects in the support set, setting a new state-of-the-art performance for weakly-supervised FSS.
\end{itemize}

\section{Related Work}

{\bf Semantic segmentation} aims to label each pixel of an image, with the category label corresponding to the content represented at that location in the image. Long et al. \cite{long2015fully} were the first to design a fully convolution network to perform this task. Due to its impressive experimental results, deep learning began to be widely used in semantic segmentation \cite{noh2015learning,zhang2018context,lin2017refinenet}. These methods require a large number of annotated images for each category. The number of labeled images for a new class is the key difference between these methods and the proposed method in this paper.

{\bf Few-shot learning} aims to learn new concepts from few examples, an ability that humans naturally possess, but machines still lack. Current FSL methods use data augmentation, model constraints and search algorithms to address this problem \cite{wang2020generalizing}. Data augmentation uses existing category information to predict the distribution of an unseen category, and then generate images for the unseen category that are used for a supervised training of a model \cite{hariharan2017low,wang2018low}. Model constraints are methods, such as fine-tuning, that allow the parameters of a network to be adapted to new categories using few examples  \cite{benaim2018one,triantafillou2017few}. Search algorithm methods project images into a feature-space and compare the features for different images to determine whether or not they belong to the same category \cite{finn2017model,grant2018recasting}. Most FSL methods focus on image classification rather than segmentation. %, while FSS is attracting substantial attention.
%Search algorithm methods searches the similarity in same category hypothesis space, to differentiate whether two images are of the same category \cite{finn2017model,grant2018recasting}. Few-shot segmentation approaches largely follow the search algorithm approach.

{\bf Few-shot segmentation} aims to perform semantic segmentation using few annotated images. Shaban et al. \cite{shaban2017one} were the first to propose a method for FSS and this approach has been the foundation for many subsequent improvements. 
%Recent efforts mostly focus on the different designs of the two-branch architecture. 
CANet \cite{zhang2019canet} designed a refinement module to enhance the feature representations by extracting object features iteratively.
BriNet \cite{yang2020brinet} encouraged more informative interactions between the extracted features from the query and support images, by emphasizing the common objects. %It also designed a multi-path strategy which was able to make better use of the support feature representations.
PMMs \cite{yang2020prototype} extracted multiple prototypes from diverse image regions to account for the diversity in appearance between different parts of the same object and different members of the same category.
%ASGNet \cite{li2021adaptive} extracted more representative prototypes by aggregating similar feature vectors to adapt to object scale and shape variation.
In addition to a segmentation loss, RePRI \cite{boudiaf2020few} used the statistics of  unlabeled pixels to add new terms to the loss function and improve performances.
%added a entropy of posteriors on the unlabeled query pixels using their statistics and it also supplemented a KL-divergence regularizer based on the proportion of the predicted foreground region to constrain the FSS model.
PFENet \cite{tian2020prior} included a training-free prior mask generation method that not only retained generalization power but also improved model performances.
However, most previous methods paid more attention to the objects, while ignoring the importance of background, negative, samples. Our proposed method is unique in using  background information from the image itself to make the model more discriminative.

%Nguyen et al. \cite{nguyen2019feature} used boosting to improve segmentation results. Boosting fuses different iterative prototypes together to predict the final results.
%In addition, many FSS methods learn from the FSL classification algorithms whose key ideas are related.
%Siam et al. \cite{siam2019adaptive} created multi-resolution proxies to improve the the extraction process. In addition, they took the imprinted weights idea from FSL classification which directly sets the final layer weights of a Convnet classifier for novel few-shot categories \cite{qi2018low}. 
%Zhang et al. \cite{zhang2019pyramid} applied graph attention \cite{velivckovic2017graph} to solve the FSS problem. They use Graph Attention Unit to combine the query graph (nodes calculated from the query set) and the support graph.

\section{Task Description}

The FSS task trains a model with $\mathcal{D}_{train}$ which is a dataset consisting of samples from class set $\mathcal{C}_{train}$, then uses this trained model to predict new classes, $\mathcal{C}_{test}$, for which only a few annotated examples are available. The key point is that $\mathcal{C}_{test} \notin \mathcal{C}_{train}$.
%meaning that the method needs to be able to segment images with access to only a few images of this class.
During training, the method has access to a large set $\mathcal{D}_{train}$ of image-mask pairs ${(I^{j}, M^{j})}^{Num}_{j=1}$, where $M^{j}$ is the semantic segmentation mask for the training image $I^{j}$, and Num is the number of image-mask pairs. During testing, the support set $S={(I^{i}_{s}, M^{i}_{s})^{k}_{i=1}}$ is a small set of k image-mask pairs, where $M^{i}_{s}$ is the semantic segmentation mask for support image $I^{i}_{s}$. A query (or test) set $Q={(I_{q}, M_{q})}$ is used to test the model, where $M_{q}$ is the ground-truth mask for image $I_{q}$. The input to the model is the support set S and the query image $I_{q}$. The output of the model is $\hat{M}_{q}$, which is the prediction of $M_{q}$. 

\section{Method}

\begin{figure*}[t]
\centering
%\fbox{\rule{0pt}{2in} %\rule{0.9\linewidth}{0pt}}
\includegraphics[width=0.75\linewidth, ]{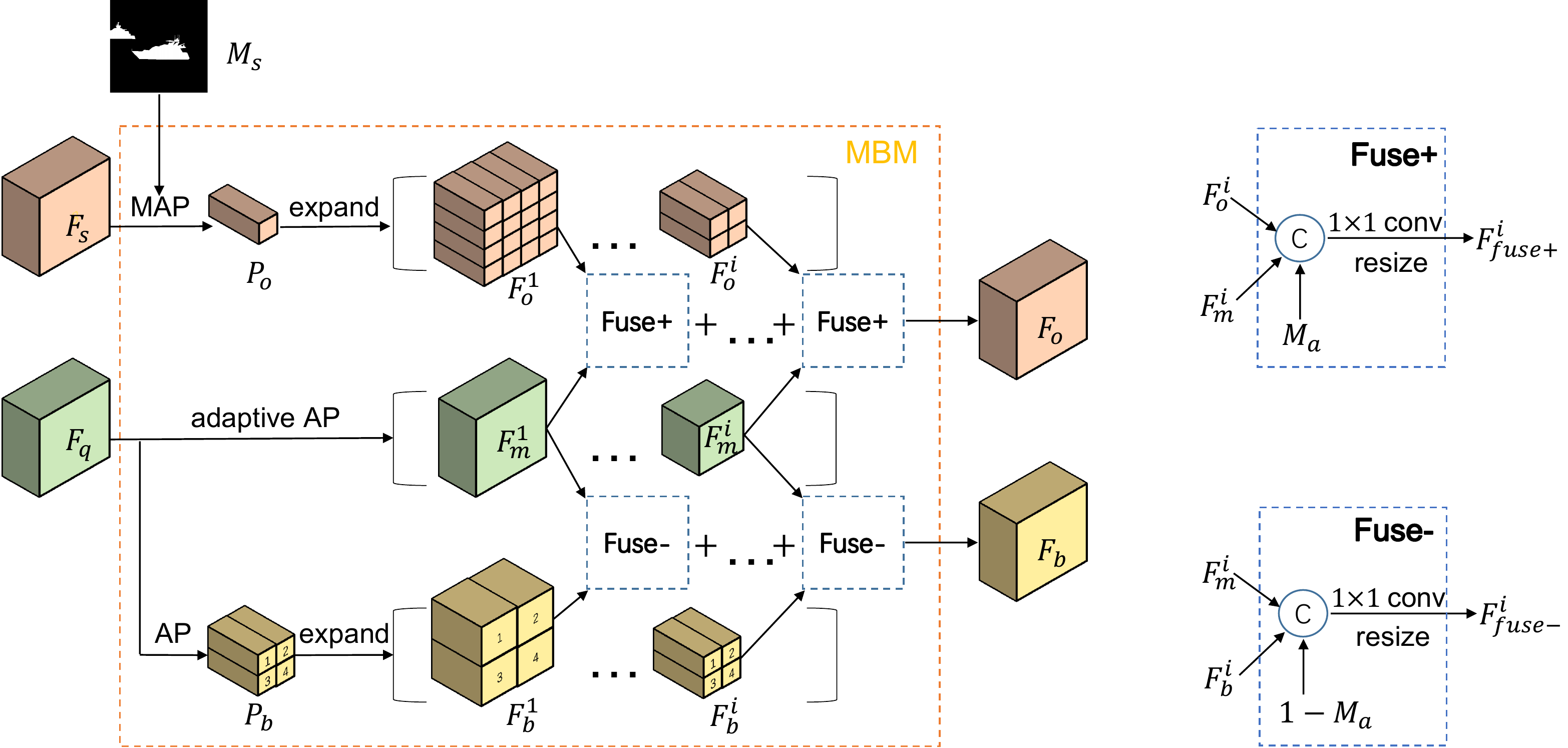}
   \caption{The multi-scale background module (MBM). The object prototype $P_{o}$ and background prototypes $P_{b}$ are extracted from the support and query features, $F_{s}$ and $F_{q}$, respectively. $P_{o}$ and $P_{b}$ are expanded to the same size as each $F_{m}^{i}$ to obtain $F_{o}^{i}$ and $F_{b}^{i}$. In the fuse module, $M_{a}$ or $1-M_{a}$ is used to activate object or background feature areas. The outputs of the fuse module are added together to get $F_{o}$ and $F_{b}$.}
\label{fig_MBMfuse}
%\end{center}
\end{figure*}

In this section, we introduce our framework, CobNet and its two constituent modules: the multi-scale background module (MBM) and the cross attention module (CAM). The overall structure of CobNet is shown in figure \ref{fig_overview}. Like other FSS methods, CobNet uses a shared ResNet-50 \cite{he2016deep} as the backbone for  feature-extraction. Following \cite{zhang2019canet,tian2020prior}, the features $F_{s}$ and $F_{q} \in \mathbb{R}^{c \times h \times w}$ extracted from the support set and query set images are the activations of the conv3\_x layers. 
%These features are the output of the conv3\_x layers of the ResNet-50.
$F_{s}$ and $F_{q}$, together with the support image segmentation mask $M_{s}$, are provided as input to the MBM.

%$F_{s}$ and its mask $M_{s}$ are feed into MBM to obtain $P_{o}$. $F_{q}$ are used to obtain multi-scale features $F_{m}$ and $P_{b}$. $F_{m}$ is combined with $P_{o}$ and $P_{b}$ respectively, to generate features $F_{o}$ and $F_{b}$ representing objects and background of the query image. 

%CAM is used to obtain a cross attention of $F_{o}$ and $F_{b}$, and predict the final result.

\subsection{Multi-scale Background Module (MBM)}
\label{sec_MBM}

The process performed by MBM is illustrated in figure \ref{fig_MBMfuse}. 
%MBM takes the support feature $F_{s}$, its mask $M_{s}$ and the query feature $F_{q}$ as inputs. 
Like other methods \cite{zhang2019canet,tian2020prior}, the prototype of the object, $P_{o} \in \mathbb{R}^{c \times 1 \times 1}$, is generated by masked average pooling (MAP) of $F_{s}$ within the masked area defined by the resized $M_{s}$.

\begin{equation}
P_{o}=\frac{\sum_{i=1}^{hw}F_{s}(i) \cdot \mathbb{1}[M_{s}(i) = 1]}{\sum_{i=1}^{hw}\mathbb{1}[M_{s}(i)=1]}
\label{eq:MAP}
\end {equation}

\noindent where $i$ indexes the spatial locations of features, and $\mathbb{1}[\cdot]$ is the indicator function, which equals 1 if the argument is True and 0 otherwise. 
For tasks in which the support set contains multiple image-mask pairs (\eg 5-shot FSS), $P_{o}$ is extracted from each support image, and then these vectors are averaged together. 

Due to the unavailability of masks for query images, the prototypes of the background, $P_{b} \in \mathbb{R}^{c \times j \times j}$, are obtained from $F_{q}$ by average pooling (AP) within $j \times j$ equal-sized, non-overlaping, regions of $F_{q}$. In figure~\ref{fig_MBMfuse}, $j=2$ is used for illustration purposes, however, $j$ is a hyper-parameter and the results of experiments to evaluate the effects of different values for $j$ are reported in Section~\ref{sec-ablation}.

%We make sure $\forall{i}, j < k_{i}$ and draw $j$ as 2 in . 

Adaptive average pooling is applied to $F_{q} \in \mathbb{R}^{c \times h \times w}$ to generate $[F_{m}^{1}, F_{m}^{2}, ..., F_{m}^{i}]$, where $F_{m}^{i} \in \mathbb{R}^{c \times k_{i} \times k_{i}}$, and $i=1, 2, ..., N$. Subscript $m$ means multi-scale and $k_{i}$ denotes the $i^{th}$ of $N$ different spatial sizes. 
$P_{o}$ and $P_{b}$ are expanded to each of these $i$ spatial sizes to generate $[F_{o}^{1}, F_{o}^{2}, ..., F_{o}^{i}]$ and $[F_{b}^{1}, F_{b}^{2}, ..., F_{b}^{i}]$, where $F_{o,b}^{i} \in \mathbb{R}^{c \times k_{i} \times k_{i}}$.  
Note that $j$ is chosen to be less than (or equal to) $k_N$ to allow $P_{b}$ to be expanded to the size of the smallest $F_{m}$.

A fuse module is designed to merge $F_{o}^{i}$ or $F_{b}^{i}$ with $F_{m}^{i}$. The specific structures of the fuse modules are shown on the right side of figure \ref{fig_MBMfuse}. Following the FSS method proposed in \cite{tian2020prior}, we use a training-free method to get an align mask $M_{a}$ to activate the feature area of interest. $M_{a}$ predicts the probability of each pixel belonging to the target object. This align mask $M_{a}$ is a prior probability mask which is obtained online, during inference, directly from the features $F_{q}$ and $F_{s}$ extracted by the backbone using:
%\vspace*{-1mm}
\begin{equation}
FM(j) = F_{s}(j) \cdot \mathbb{1}[M_{s}(j)=1] \; j \in \{1,2,..,hw\}
\end {equation}

\begin{equation}
M_{a} = \mathop{max}\limits_{j \in \{1,2,...,hw\}} (cos(F_{q}(i),FM(j)) \; i \in \{1,2,..,hw\})
\end {equation}

\noindent where $cos$ is the cosine similarity. $M_{a}$ is down-sampled using  bilinear interpolation to fit the shape of $F_{m}^{i} \in \mathbb{R}^{1 \times k_{i} \times k_{i}}$.

The Fuse+ module concatenates $M_{a}$ with $F^{i}_{o}$ and $F^{i}_{m}$ while the Fuse- module concatenates $1-M_{a}$ with $F^{i}_{m}$ and $F^{i}_{b}$. This is followed by a layer of $1 \times 1$ convolution. Fuse+ and Fuse- learn to extract features that are predictive of objects and background respectively. The output of each fuse module, $F^{i}_{fuse+}$ or $F^{i}_{fuse-}$, is generated by down-sampling the number of channels to c and resizing its shape to $h \times w$. All outputs of the fuse modules at different scales are added together to get the final output $F_{o}$ or $F_{b}$ (where $F_{o, b} \in \mathbb{R}^{c \times h \times w}$).

%\begin {equation}
%F_{o, b} = \sum_{i=1}^{N} \textbf{Fuse}(F_{o, b}^{i}, F_{m}^{i})
%\label{Fo,b}
%\end {equation}

%Where $F_{o}^{i}$, $F_{b}^{i}$ and $F_{m}^{i}$ respectively represents,

%\begin {equation}
%F_{o}^{i} = \mathrm{EP}^{i}(\mathrm{GMP}(F_{s}, mask)), F_{b}^{i} = \mathrm{EP}^{i}(\mathrm{Pool}(F_{q})), F_{m}^{i} = \mathrm{AP}^{i}(F_{q})
%\label{Fn}
%\end {equation}

%Here $\mathrm{EP}^{i}$ means expanding $P_{o,b}$ to features $F_{o,b}^{i}$. $\mathrm{Pool}$ and $\mathrm{AP}^{i}$ respectively represent pooling and adaptive pooling.

\subsection{Cross Attention Module (CAM)}
\label{sec_CAM}

\begin{figure}[t]
\centering
%\fbox{\rule{0pt}{2in} %\rule{0.9\linewidth}{0pt}}
\includegraphics[width=0.8\linewidth]{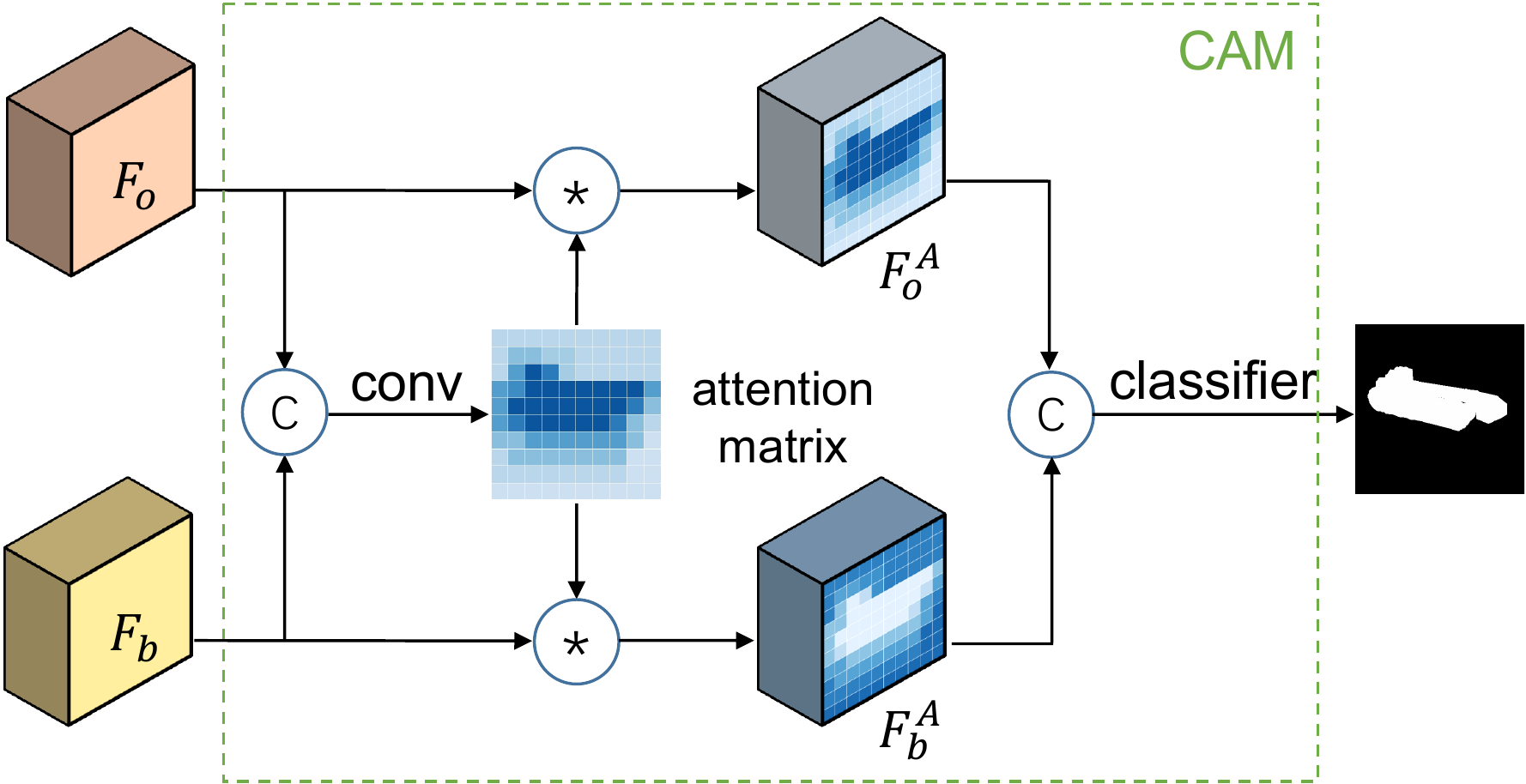}
   \caption{The cross attention module (CAM). $F_{o}$ and $F_{b}$ are concatenated together, and then passed through two convolution layers to generate an attention matrix. This matrix is used as a weight to activate $F_{o}$ and $F_{b}$. Finally, a classification head is used to get a final prediction mask.}
\label{fig_CAM}
%\end{center}
\vspace{-1mm}
\end{figure}

Because the background features ($F_{b}$) have been obtained in an unsupervised way, they inevitably include information from the objects in the query image. CAM is designed to activate the more accurate parts of object and background features. Figure \ref{fig_CAM} shows the details of CAM.  We first concatenate $F_{o}$ and $F_{b}$ together, to get a feature $(\mathbb{R}^{2c \times h \times w})$. After 2 layers of $1 \times 1$ convolution
and a sigmoid layer, a cross attention matrix, $A$, is generated with a size of $1 \times h \times w$. This attention matrix is used to weight the foreground and background features, as follows:
\begin {equation}
F_{o}^{A} = F_{o} \circledast A, \;\; \; \; \; \; \; \;  F_{b}^{A} = F_{b} \circledast (1-A)
\label{Fatte}
\end {equation}
Where $\circledast$ means element-wise multiplication. Compared to concatenation of the align mask (in figure \ref{fig_MBMfuse}), the operation of element-wise multiplication has a greater impact on features. A trainable attention matrix can activate the valuable areas of each input features, object information in $F_{o}$ and background information in $F_{b}$, alleviating their respective inaccurate areas.

%\subsection{Training}

$F_{o}^{A}$ and $F_{b}^{A}$ are concatenated together and provided as input to the classifier. The classifier includes 3 layers of $3 \times 3$ convolution and a layer of $1 \times 1$ convolution, which is used to get a final prediction mask. We calculate a cross-entropy loss $\mathcal{L}_{seg}$ for this final prediction mask. Following \cite{tian2020prior}, losses are also calculated for the intermediate results produced by the MBM. With two layers of $3 \times 3$ convolution, each $F^{i}_{fuse+}$ (figure \ref{fig_MBMfuse}) is used to create a prediction mask. We calculate $N$ cross-entropy losses $\mathcal{L}_{i} (i = {1, 2,.., N})$ from these prediction masks. The total loss is:
\begin {equation}
\mathcal{L} = \frac{1}{N} \sum_{i=1}^{N} \mathcal{L}_{i} + \mathcal{L}_{seg}
\label{loss}
\end {equation}

\subsection{Generalization to weaker annotations}

FSS, while only using a limited amount of supervision (\ie 1 or 5-shots), uses strong supervision (in the form of a pixel-wise segmentation mask) for the few examples in the support set. It is labour intensive to provide such annotations for every novel class, and may be impractical in some applications.
To address this issue, previous weakly-supervised FSS methods (while using segmentation masks during training) have experimented with replacing the pixel-wise segmentation masks (in the support set during testing) with less precise information, such as bounding boxes \cite{wang2019panet,zhang2019canet}, or image-level object category labels \cite{siam2020weakly}.
%Our model is generalized and is able to be directly applicable to weaker annotations.
We evaluate CobNet with an even more extreme form of weak-supervision by providing, at test time, no annotations of the support images at all. In this case, the object prototype ($P_{o}$ in figure \ref{fig_MBMfuse}) is calculated using global AP rather than MAP. In addition, as a mask is required to generate the align mask ($M_{a}$ in figure 2), we set the mask to a matrix of all ones. 
%Note, that in the current work, as in all previous work on weakly-supervised FSS \cite{wang2019panet,zhang2019canet,siam2020weakly}, the segmentation masks of the training images are still used to calculate the loss during training.

%Panet and Canet obtain prototypes using segmentation mask during training, but using bounding box during testing. 
%TOSFL obtains prototypes by image-level during training and testing.

\section{Experiments}
\subsection{Experimental Setup}

\subsubsection{Datasets} The PASCAL-$5^{i}$ \cite{shaban2017one} and COCO-$20^{i}$ \cite{nguyen2019feature} datasets were used for evaluation. PASCAL-$5^{i}$ includes the PASCAL VOC2012 \cite{everingham2010PASCAL} and the extended SDS datasets \cite{hariharan2014simultaneous}. It contains 20 classes which are divided into 4 folds each containing 5 classes. COCO-$20^{i}$ is the MS-COCO dataset \cite{lin2014microsoft} with the 80 classes divided into 4 folds each containing 20 classes. Performance was measured using 4-fold cross-validation: testing each fold in turn using a model that had been trained on the other three folds. For each fold, testing was performed using a random sample of 1,000 query-support pairs.

\subsubsection{Implementation details} 
%We used a ResNet-50 \cite{he2016deep} as the backbone for extracting features from the input images. 
Input images were resized to $473 \times 473$ pixels. The parameters of the ResNet-50 \cite{he2016deep} backbone were initialized from a model pre-trained on ImageNet \cite{russakovsky2015imagenet} and were kept fixed during FSS training. The other parameters in CobNet were trained by SGD, with a momentum of 0.9, for 200 epochs on PASCAL-$5^{i}$ and 50 epochs on COCO-$20^{i}$. The learning rate was initialized to 0.001 and reduced following the “poly” policy \cite{chen2017deeplab}. $j = 4$ (see section~\ref{sec_MBM}) was used in all experiments except those whose results are shown in table~\ref{tab_num}. N was set to 4 and $k_{i}$ was set to [60, 32, 16, 8]. Our approach was implemented in PyTorch and run on Nvidia Tesla V100 GPUs. The batch size was set to 4 during training. Horizontal flip and random rotation were used as data augmentation methods.

\subsubsection{Evaluation metric} Mean intersection over union (mIoU) was used as the main evaluation metric. It computes the IoU for each individual foreground class and then averages these values over all classes (5 in PASCAL-$5^{i}$ and 20 in COCO-$20^{i}$). We also report the results of FB-IoU, which calculates the mean IoU for the foreground (\ie for all objects ignoring class labels) and the background.

\begin{table*}[ht]
\centering
\caption{mIoU (\%) results for 1-shot and 5-shot FSS on the four folds of COCO-$20^{i}$. ‘Mean’ is the mIoU averaged across folds. The best result for each column is in bold. Methods listed above the horizontal line use VGG16~\cite{simonyan2014very} for feature extraction while those below the line use ResNet-50~\cite{he2016deep}.}
\label{coco_all}
\begin{tabular}{lllllllllll}
\hline
           & \multicolumn{5}{c}{1 shot}         & \multicolumn{5}{c}{5 shot}          \\ \cmidrule(r){2-6} \cmidrule(r){7-11} 
Method & C-$20^{0}$ & C-$20^{1}$ & C-$20^{2}$ & C-$20^{3}$ & \hspace*{-1ex}Mean\hspace*{-1ex} & C-$20^{0}$\ & C-$20^{1}$ & C-$20^{2}$ & C-$20^{3}$ & Mean \\ \hline
FWB \cite{nguyen2019feature} & 18.4 & 16.7 & 19.6 & 25.4 & 20.0 & 20.9 & 19.2 & 21.9 & 28.4 & 22.6 \\
PANet \cite{wang2019panet} & - & - & - & - & 20.9 & - & - & - & - & 29.7 \\ \hline
RPMMs \cite{yang2020prototype} & 29.5 & 36.8 & 28.9 & 27.0 & 30.6 & 33.8 & 42.0 & 33.0 & 33.3 & 35.5\\
PFENet \cite{tian2020prior}  & 34.3 & 33.0 & 32.3 & 30.1 & 32.4 & 38.5 & 38.6 & 38.2 & 34.3 & 37.4\\
ASR \cite{liu2021anti} & 29.9 & 35.0 & 31.9 & 33.5 & 32.6 & 31.3 & 37.9 & 33.5 & 35.2 & 34.4\\
CWT \cite{lu2021simpler} & 32.2 & 36.0 & 31.6 & 31.6 & 32.9 & 40.1 & 43.8 & 39.0 & 42.4 & 41.3\\
RePRI \cite{boudiaf2020few}  & 32.0 & 38.7 & 32.7 & 33.1 & 34.1 & 39.3 & 45.4 & 39.7 & \textbf{41.8} & 41.6\\
ASGNet \cite{li2021adaptive}  & \textbf{34.9} & 36.9 & 34.3 & 32.1 & 34.6 & \textbf{41.0} & \textbf{48.3} & 40.1 & 40.5 & \textbf{42.5}\\
CobNet (ours)\hspace*{-2mm} & 34.5 & \textbf{42.4} & \textbf{36.7} & \textbf{37.8} & \textbf{37.8} & 34.8 & 44.6 & \textbf{42.2} & 39.1 & 40.2 \\ \hline
\end{tabular}
\label{tab_COCO}
%\vspace{-1mm}
\end{table*}

\begin{table*}[tbp]
\centering
\caption{mIoU (\%) results for 1-shot and 5-shot FSS on the four folds of PASCAL-$5^{i}$. Otherwise the format of this table is identical to, and described in the caption of, Table \ref{tab_COCO}.}
\begin{tabular}{lllllllllll}
\hline
           & \multicolumn{5}{c}{1-shot}         & \multicolumn{5}{c}{5-shot}         \\ \cmidrule(r){2-6} \cmidrule(r){7-11}
Method & P-$5^{0}$ & P-$5^{1}$ & P-$5^{2}$ & P-$5^{3}$ & Mean & P-$5^{0}$ & P-$5^{1}$ & P-$5^{2}$ & P-$5^{3}$ & Mean \\ \hline
%OSLSM \cite{shaban2017one} & 33.6 & 55.3 & 40.9 & 33.5 & 40.8 & 35.9 & 58.1 & 42.7 & 39.1 & 43.9 \\
FWB \cite{nguyen2019feature} & 47.0 & 59.6 & 52.6 & 48.3 & 51.9 & 50.9 & 62.9 & 56.5 & 50.1 & 55.1\\
PANet \cite{wang2019panet} & 42.3 & 58.0 & 51.1 & 41.2 & 48.1 & 51.8 & 64.6 & 59.8 & 46.5 & 55.7\\ \hline
CANet \cite{zhang2019canet} & 52.5 & 65.9 & 51.3 & 51.9 & 55.4 & 55.5 & 67.8 & 51.9 & 53.2 & 57.1\\
%SimProp \cite{gairola2020simpropnet}& 54.9 & 67.3 & 54.5 & 52.0 & 57.2 & 57.2 & 68.5 & 58.4 & 56.1 & 60.0\\
RPMMs \cite{yang2020prototype}& 55.2 & 66.9 & 52.6 & 50.7 & 56.3 & 56.3 & 67.3 & 54.5 & 51.0 & 57.3\\
PFENet \cite{tian2020prior}  & 61.7 & 69.5 & 55.4 & 56.3 & 60.8 & 63.1 & 70.7 & 55.8 & 57.9 & 61.9\\
ASR \cite{liu2021anti} & 53.8 & 69.6 & 51.6 & 52.8 & 56.9 & 56.2 & 70.6 & 53.9 & 53.4 & 58.5 \\
CWT \cite{lu2021simpler} & 56.3 & 62.0 & 59.9 & 47.2 & 56.4 & 61.3 & 68.5 & 68.5 & 56.6 & 63.7\\
RePRI \cite{boudiaf2020few} & 59.8 & 68.3 & \textbf{62.1} & 48.5 & 59.7 & \textbf{64.6} & \textbf{71.4} & \textbf{71.1} & 59.3 & \textbf{66.6}\\
ASGNet \cite{li2021adaptive} & 58.8 & 67.9 & 56.8 & 53.7 & 59.3 & 63.7 & 70.6 & 64.2 & 57.4 & 63.9\\
CobNet (ours)  & \textbf{62.7} & \textbf{69.7} & 55.1 & \textbf{58.0} & \textbf{61.4} & 64.0 & \textbf{71.4} & 55.3 & \textbf{60.9} & 62.9 \\ \hline
\end{tabular}
\label{tab_all}
%\vspace{-1mm}
\end{table*}

\begin{table}[tbp]
 \centering
\caption{FB-IoU (\%) results for 1-shot and 5-shot FSS on PASCAL-$5^{i}$.  'Params' is the  number of learnable parameters.}
\begin{tabular}{llll}        \hline
Method & 1-shot & 5-shot & Params  \\ \hline
OSLSM \cite{shaban2017one}  & 61.3 & 61.5 & 276.7M \\ 
PANet \cite{wang2019panet}  & 66.5 & 70.7 & 14.7M  \\ \hline
CANet \cite{zhang2019canet} & 66.2 & 69.6 & 19.0M  \\
%SimProp \cite{gairola2020simpropnet} & 73.0 & 72.9 & -  \\
%RPMMs \cite{yang2020prototype} & - & - & - \\
%RePRI \cite{boudiaf2020few} & - & - & -  \\
PFENet \cite{tian2020prior}  & \textbf{73.3} & 73.9 & 10.8M \\
ASR \cite{liu2021anti} & 71.3 & 72.5 & - \\
ASGNet \cite{li2021adaptive}  & 69.2 & 74.2 & \textbf{10.4}M  \\ 
CobNet (ours) & 72.4 & \textbf{74.4} & 10.8M  \\\hline
      \end{tabular}
  \label{tab_P-FB}
\vspace{-1mm}
\end{table}

\subsection{Results}

\subsubsection{COCO-$20^{i}$} Table \ref{tab_COCO} compares the performance of various methods on COCO-$20^{i}$. CobNet achieves new state-of-the-art results in the 1-shot setting. Compared to the second best method, ASGNet, CobNet improves the mean mIoU by around 3.2\%. Compared with PASCAL-$5^{i}$, COCO-$20^{i}$ is a more challenging dataset with diverse samples and categories. These results, therefore, demonstrate that CobNet is able to handle complex images, and supports our claim that it is advantageous to extract background information from the query image itself.

\subsubsection{PASCAL-$5^{i}$} Tables \ref{tab_all} and \ref{tab_P-FB} compares the performance of CobNet with recent state-of-art methods.  In terms of mIoU, CobNet outperforms other methods in the 1-shot setting and achieves the best results on the majority of folds across both 1-shot and 5-shot segmentation. In terms of FB-IoU, CobNet produces results that are competitive with the state-of-the-art for the 1-shot setting, and that improve on previous performances in 5-shot segmentation. In addition, the number of trainable parameters is lower than, or similar to, other methods.

\subsubsection{Weakly-supervised} Despite using even weaker supervision than has been used for previous algorithms, CobNet achieves state-of-the-art results as shown in Table \ref{tab_nomask}. These results demonstrate that the proposed method is capable of handling noise in both the object and background information. 
%Without any labeling, CobNet is still able to maintain a certain effect.

\begin{table}[tbp]
\caption{mIoU (\%) results for 1-shot weakly-supervised FSS on PASCAL-$5^i$. Methods vary in the type of supervision provided: `BB' indicates the use of bounding boxes, `IL' indicates the use of image-level labels.}
\centering
\begin{tabular}{lcclllll}
\hline
Method & \multicolumn{2}{c}{Supervision} & P-$5^{0}$ & P-$5^{1}$ & P-$5^{2}$ & P-$5^{3}$ & Mean \\ 
& BB & IL \\ \hline
Panet \cite{wang2019panet} & yes & no & - & - & - & - & 45.1 \\ 
Canet \cite{zhang2019canet} & yes & no & - & - & - & - & 52.0 \\ 
TOSFL \cite{siam2020weakly} & no & yes & 42.5 & \textbf{64.8} & 48.1 & \textbf{46.5} & 50.5 \\ 
CobNet (ours) & no & no & \textbf{56.3} & 63.1 & \textbf{48.9} & \textbf{46.5} & \textbf{53.7} \\ \hline
\end{tabular}
\label{tab_nomask}
\vspace{-1mm}
\end{table}

\begin{figure*}[tb]
\centering
%\fbox{\rule{0pt}{2in} \rule{.9\linewidth}{0pt}}
\includegraphics[width=\textwidth]{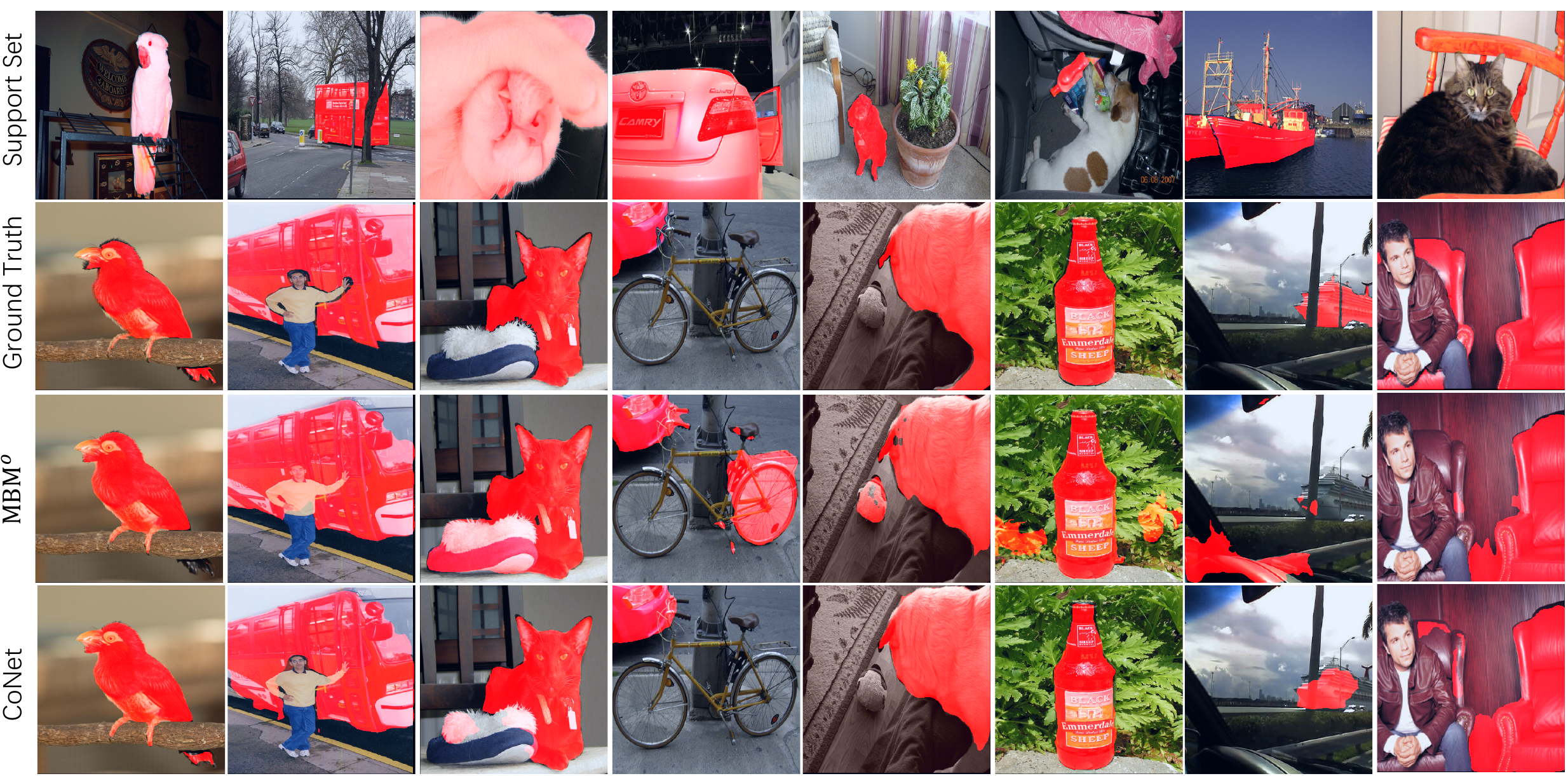}
   \caption{Qualitative results of 1-shot FSS on PASCAL-$5^{i}$. Best viewed in color.}
\label{fig_qualitative}
%\end{center}
%\vspace{-1mm}
\end{figure*}

\subsection{Ablation Study}
\label{sec-ablation}

%Extensive ablation studies, using PASCAL-$5^i$, were performed to verify the effectiveness of the proposed method.

\subsubsection{Impact of each module} Table \ref{tab_ablation} shows the impact of each module. For models in which the CAM module is ablated, the MBM directly feeds the concatenation of its output features into a classifier (which has the same architecture as the classifier described in section~\ref{sec_CAM}) to predict the segmentation mask. Different, ablated versions of the MBM are used. MBM$^o$ refers to a model in which the MBM only extracts object features and no background features. MBM$^{s}$ extracts the background prototypes from the support set by MAP rather than from the query images using AP. 

It can be seen that extracting background information improves performances, and that when background information is extracted from the support images it causes the greatest improvement. The CAM is able to exploit this background information to improve results further, especially when the background prototypes come from the query image.

\begin{table}[tb]
  \centering
\caption{Ablation Study. Mean mIoU (\%) results for 1-shot FSS on PASCAL-$5^i$ and COCO-$20^{i}$. MBM$^{o}$ is an ablated version of MBM which only extracts object features without background features. MBM$^{s}$ is an ablated version of MBM which obtains background prototypes from the support set by MAP.}
%\caption{Ablation Study. Mean mIoU (\%) results of 1-shot FSS on PASCAL-$5^i$. MBM$^o$ is an ablated version of MBM which only extracts foreground features (and not background features) from the input images. MBM$^{s}$ is an ablated version of MBM in which background prototypes were obtained from the support set by MAP (rather than from the query images using AP). }
\begin{tabular}{lll}        \hline
Method & \multirow{2}{20mm}{PASCAL-$5^i$ Mean} & \multirow{2}{20mm}{COCO-$20^i$ Mean}\\
\\\hline
MBM$^o$  & 59.2 & 33.6 \\ 
MBM$^{s}$ & 59.6 & 34.5 \\
MBM  & 60.8 & 35.8 \\ 
MBM$^{s}$\, + CAM  & 60.9 & 36.4 \\
MBM \; + CAM & \textbf{61.4} & \textbf{37.8} \\ \hline
\end{tabular}
%\vspace{-1mm}
\label{tab_ablation}
\end{table}

\subsubsection{Number of background prototypes} Table \ref{tab_num} shows the effects of varying the number of background prototypes initially extracted by the MBM. When $j=4$, CobNet provides the best experimental result. It can also be seen that when $j$ is either too small, or too large, the performance is harmed. However, in all cases the performance is competitive to previous state-of-the-art results (\cf Table~\ref{tab_all}). The weakened performance is likely to be due to too much, or to too little, smoothing when averaging feature-vectors across large or small regions of the image to produce the background prototypes. Specifically, average pooling over the whole image (when $j=1$) is likely to result in prototypes unable to express rich and diverse background information. While average pooling in only small parts of the image (when $j=8$) is likely to produce prototypes that are idiosyncratic and unrepresentative. 

\begin{table}[ht]
\centering
\caption{Effect of the number of background prototypes. Mean mIoU (\%) results for 1-shot FSS on PASCAL-$5^{i}$. $j \times j$ determines the number of background prototypes, as described in section~\ref{sec_MBM}}
\begin{tabular}{ll}   \hline
$j \times j$  & Mean \\ \hline
$1 \times 1 = 1$  & 60.6 \\ 
$2 \times 2 = 4$ & 61.1 \\ 
$4 \times 4 = 16$ & \textbf{61.4} \\ 
$8 \times 8 = 64$  & 60.6 \\ \hline
\end{tabular}
\label{tab_num}
%\vspace{-1mm}
\end{table}

\subsubsection{Qualitative results} Figure \ref{fig_qualitative} shows some examples of 1-shot segmentation results. It can be seen that the proposed CobNet produces accurate semantic segmentation predictions even when there are large variations in the appearance of the objects and background between the support and query images. On the right side of figure \ref{fig_qualitative}, is an example of segmentation failure as the left chair is segmented incompletely, and part of the wall is mis-classified as chair. This illustrates a limitation of our simple method for extracting background information from the query image which has not helped in this example, where the appearances of the objects (red chairs) and the background (red wall) are similar.

\section{Conclusion}

This paper has introduced CobNet as a new architecture for few-shot image
segmentation. The proposed architecture extracts information about the appearance of background features from the query image itself. Our experimental results demonstrate that this approach produces state-of-the-art performances in 1-shot FSS, and weakly supervised FSS, when evaluated using standard benchmarks. Furthermore, ablation experiments demonstrate the advantages of using the query image as the source of background information. Exploring methods to more accurately extract background features from the query image may be a promising future direction for further improving FSS.

% conference papers do not normally have an appendix

% use section* for acknowledgment
%\section*{Acknowledgment}

%The authors would like to thank...

% trigger a \newpage just before the given reference
% number - used to balance the columns on the last page
% adjust value as needed - may need to be readjusted if
% the document is modified later
%\IEEEtriggeratref{8}
% The "triggered" command can be changed if desired:
%\IEEEtriggercmd{\enlargethispage{-5in}}

% references section

% can use a bibliography generated by BibTeX as a .bbl file
% BibTeX documentation can be easily obtained at:
% http://mirror.ctan.org/biblio/bibtex/contrib/doc/
% The IEEEtran BibTeX style support page is at:
% http://www.michaelshell.org/tex/ieeetran/bibtex/
%\bibliographystyle{IEEEtran}
% argument is your BibTeX string definitions and bibliography database(s)
%\bibliography{IEEEabrv,../bib/paper}
%
% <OR> manually copy in the resultant .bbl file
% set second argument of \begin to the number of references
% (used to reserve space for the reference number labels box)

{\small
\bibliographystyle{IEEEtran}
\bibliography{IEEEfull}
}

% that's all folks
\end{document}